\documentclass[10pt, a4paper]{article}

\usepackage[final]{lrec-coling2024} 

\title{Is Spoken Hungarian Low-resource?: \\A Quantitative Survey of Hungarian Speech Data Sets}

\name{Péter Mihajlik$^{\ast, \dagger}$, Katalin Mády$^{\ast}$, Anna Kohári$^{\ast}$, Fruzsina Sára Vargha$^{\ast}$,
\\ {\bf \large Gábor Kiss$^{\dagger}$, Tekla Etelka Gráczi$^{\ast}$,  A. Seza Doğruöz$^{\ddag}$}  }

\address{\\ $^{\ast}$HUN-REN Hungarian Research Centre for Linguistics \\ 
{mihajlik.peter, mady.katalin, kohari.anna, graczi.tekla.etelka}@nytud.hu \\ \\$^{\dagger}$Budapest University of Technology and Economics 
\\ Departure of Telecommunications and Media Informatics \\ 
kiss.gabor@tmit.bme.hu \\ \\
$^{\ddag}$LT3, IDLab, Universiteit Gent\\ as.dogruoz@ugent.be}

\abstract{
Even though various speech data sets are available in Hungarian, there is a lack of a general overview about their types and sizes. To fill in this gap, we provide a survey of available data sets in spoken Hungarian in five categories (e.g., monolingual, Hungarian part of multilingual, pathological, child-related and dialectal collections). In total, the estimated size of available data is about 2800 hours (across 7500 speakers) and it represents a rich spoken language diversity. However, the distribution of the data and its alignment to real-life (e.g. speech recognition) tasks is far from optimal indicating the need for additional larger-scale natural language speech data sets. Our survey presents an overview of available data sets for Hungarian explaining their strengths and weaknesses which is useful for researchers working on Hungarian across disciplines. In addition, our survey serves as a starting point towards a unified foundational speech model specific to Hungarian.
 \\ \newline  }

\begin{document}

\maketitleabstract

\section{Introduction}

Although (written) Hungarian is one of the most used 25 languages across websites\footnote{ \url{https://w3techs.com/technologies/overview/content_language}}, annotated spoken language data are relatively scarce. Hungarian speech data with time-aligned transcriptions is difficult to find in a quantity sufficient to train a contemporary Automatic Speech Recognition (ASR).
Recent advancements in Self-Supervised Learning (SSL) \citep{baevski2020wav2vec} allow for high accuracy speech models using limited supervised (transcribed) speech data (for fine-tuning), but state-of-the-art ASR systems still use several thousand hours of transcribed speech \citep{radford2022robust}. In our survey, we provide a comprehensive quantitative overview of Hungarian labeled speech data sets accessible (at least) for research purposes. The amount of available data 
seems impressive both in terms of variability and quantity if compared to popular English ASR benchmark data sets such as \citep{switchboard} or \citep{librispeech}. However, unifying the annotations/transcriptions for a potential foundational speech model poses significant challenges. Nevertheless, we believe that such an overview is useful not only for the Speech Community and for an interdisciplinary LREC audience, but it will also pave the road to a unified Hungarian data set/speech model. 

\section{Related work}

There is a wide range of data available on the Internet for English, but not much is known about the size and types of data sources available for low and mid resource languages. Therefore, there is a need for surveys and/or overview articles on less resourced languages (e.g., \newcite{eszter2012hungarian}, \newcite{jelencsik2023language} for Hungarian and \newcite{coltekin2023} for Turkish, \newcite{rehm2014update} for several languages). 
A quantitative overview of a given language's speech resources (e.g., the number of speakers, database length in hours) is uncommon even if it is highly appreciated by researchers/developers performing statistical analysis and/or model development. This also applies to Hungarian speech corpora. Although language resource catalogs like ELRA\footnote{ \url{https://catalog.elra.info/}}, LDC\footnote{ \url{https://catalog.ldc.upenn.edu/}}, 
ELG\footnote{ \url{https://live.european-language-grid.eu/}} are helpful for hosting the desired speech data, they do not provide a quantitative overview, partly because the data owners may not have provided it when submitting their data. Another issue is that the links for data sources often become invalid over time without any contact information (cf. \cite{coltekin2023}).



\section{Characteristics of the Hungarian Language}

Hungarian belongs to the Uralic language family along with Finnish and Estonian, and it is spoken by approximately 12.5 million native speakers (mostly in Hungary, but also in neighboring countries, Europe and overseas \citep{kenesei2002hungarian}).
It is an agglutinating language with rich morpho(phono)logy subject to vowel harmony (i.e. grammatical relations are expressed mostly by suffixes\citep{siptar2000phonology}). An extended version of the Latin alphabet is used in orthography, including diacritics for vowel quantity and quality (e.g., \'o, \"o, \H{o}). It has relatively flexible word order \citep{kenesei2002hungarian}, and sentences carry stronger prominence in their initial parts. Word stress is fixed to the first syllable \citep{reichel2018acoustic}.

\section{Hungarian Speech Data Sets} 

In this section, we focus on various Hungarian speech corpora by types and consider only Hungarian audio data (accompanied at least by text transcriptions in Hungarian). In Tables \ref{tab:mono}--\ref{tab:dialect} the key features (e.g., size in terms of hours and speakers) of the data sets are summarized 

\subsection{Monolingual Hungarian Data Sets}
Table~\ref{tab:mono} provides an overview of the recordings with healthy adult speakers of Standard Hungarian. MTBA was recorded via a public switched telephone network with reduced bandwidth, a simplified, SpeechDat(E)-like (\citetlanguageresource{SPEECHDAT-E}) corpus. A larger data set is BEA (Spoken Hungarian Database) (\citetlanguageresource{gosy2012,neuberger2014}) with 472 recorded speakers, for which annotations are still in progress. The subset BEA-Base \citetlanguageresource{mihajlik_etal_lrec2022} is devoted to benchmarking ASR systems. BUSZI (Budapest Sociolinguistic Interviews) has limited access (i.e., written data can be searched online, 
but the sound files are available only on-site).

\begin{table*}[ht!]
 \centering
 \begin{tabular}{| p{3.5cm}||c|c|p{8cm} |}
 \hline
 \textbf{Name}& \textbf{[hours]}& \textbf{Speakers}& \textbf{Key Features}\\ \hline 
 BUSZI & \textasciitilde600 & 250 & Sociolinguistic Interviews recorded with magnetofon. 
 \citetlanguageresource{BUSZI}\\ 
 \hline
 Oasis Numbers & 3 & 26 & Domain-specific corpus of numbers, 5857 recorded words.
 \citetlanguageresource{OASIS}\\
 \hline
 MTBA& \textasciitilde5& 500 & Mostly read telephone speech.
 \citetlanguageresource{MTBA}\\
 \hline
 MRBA& \textasciitilde6& 332 &Phonetically balanced read text.
 \citetlanguageresource{MRBA}\\
 \hline
 BME Broadcast & 3& \textasciitilde10& TV broadcast news.
 \citetlanguageresource{Broadcast_BME}\\ 
 \hline
 Szeged Broadcast& 28& n.a.&News broadcast from 8 tv channels.
 \citetlanguageresource{Broadcast_Szeged}\\
 \hline
 HuComTech& \textasciitilde50 & 112 & Audioviusal, conversational and read speech of students.
 \citetlanguageresource{HUCOMTECH}\\
 \hline
 BEA & $~$ \textasciitilde350& 472 & Studio, various speech types: conversational, monolog, read, repeated, etc.
 \citetlanguageresource{gosy2012}\\
 \hline
 Kivi & 1 & 45 & Short guided monologues. 
 \citetlanguageresource{KIVI}\\
 \hline
 SzöSzi & $~$370& 163& Sociolinguistic interviews with Szegedian speakers, partially transcribed. \citetlanguageresource{kontra2016} \\\hline
 BEKK& 20& 56 & Spontaneous conversations of students recorded in dormitories by smartphones. 
 \citetlanguageresource{BEKK} \\
 \hline
 HuTongue& \textasciitilde500& 15& Conversations of reality show characters recorded with head microphones.
 \citetlanguageresource{HUTONGUE}\\
 \hline
 StaffTalk& 101& 20& Spontaneous conversations of teachers recorded through smartwatches. 
 \citetlanguageresource{STAFFTALK} \\
 \hline
 Akaka Maptask  \newline Corpus& 5& 46& Task-oriented dialogues recorded via head-mounted microphones.
 \citetlanguageresource{molnar_akaka2023}\\
 \hline
 Budapest Games  \newline Corpus& 9& 12& 36 task-oriented dialogues recorded via head-mounted microphones.
 \citetlanguageresource{bcg2023}\\
 \hline
 ForVoice120+& ~32& 120& Various speech tasks for speaker identification. \newline
 \citetlanguageresource{ForVoice120+}\\
 \hline
\textbf{Total approximately} & \textbf{2083} & \textbf{2179} & \\\hline
 

 \end{tabular}
 \caption{Monolingual Hungarian Data Sets}
 \label{tab:mono}
\end{table*}

\subsection{Hungarian Data Sets as Parts of Multilingual Databases}
Hungarian subsets of multilingual data sets are summarized in Table~\ref{tab:multi}. Only data with a minimum length of one hour are listed. This collection is diverse in terms of content. CommonVoice Hungarian showed a rapid growth in 2023 from 20 to 153 hours. SPEECON and SpeechDat(E) are important spoken language resources, but they lack spontaneous speech. The final data set is called HUN\_ASR001\_CN which looks promising (in terms of size and speakers), but no further details are available. 



\begin{table*}[ht]
 \centering
 \begin{tabular}{| p{3.5cm}||c|c|p{8cm} |}
 \hline
 \textbf{Name}& \textbf{[hours]}& \textbf{Speakers}& \textbf{Key Features}\\
 \hline
 BABEL& 1& 72& Clear speech, read text.
 \citetlanguageresource{BABEL}\\
 \hline
 SpeechDat(E)& 65 & 1000& Read telephone speech, ELRA.
 \citetlanguageresource{SPEECHDAT-E}\\
 \hline
 SPEECON-1& \textasciitilde200& 555& Read speech recorded through 4 microphone channels, ELRA.
 \citetlanguageresource{Speecon}\\
 \hline
 CSLU: 22 Languages & \textasciitilde4 & 300 & Prompted telephone speech, LDC.
 \citetlanguageresource{CSLU-22}\\
 \hline
 CSS10& 10& 1& Free audiobook.
 \citetlanguageresource{CSS-10}\\
 \hline
 CommonVoice& 151& 1603& Read sentences, online collection.
 \citetlanguageresource{commonvoice}\\
 \hline
 MaSS & \textasciitilde21& 1-25& Sentence-aligned Spoken Utterances Extracted from the Bible. 
 \citetlanguageresource{mass}\\
 \hline
 VoxPopuli& 63& 143& EU parliamentary speech.
 \citetlanguageresource{voxpopuli}\\
 \hline
 FLEURS& \textasciitilde12& n.a.& Read sentences from wikipedia, multi-purpose.
 \citetlanguageresource{fleurs}\\
 \hline
 HUN\_ASR001\_CN & 286 & 254 & Scripted speech data recorded via smartphones.
 \citetlanguageresource{HUN_ASR001_CN}\\
 \hline
 \textbf{Total approximately} & \textbf{813} & \textbf{3928} & \\\hline
 \end{tabular}
 \caption{Hungarian Data Sets as Parts of Multilingual Collections}
 \label{tab:multi}
\end{table*}

\subsection{Pathological Speech}
Table \ref{tab:patho} collects data sets created to support research and diagnostics related to neurological, physiological or psychiatric disorders affecting speech.
However, the speech tasks to be performed by the speakers (introduction, readings, sustained vowels etc.) are partly different across data sets which makes it difficult to examine the disorders comparatively.


\begin{table*}[ht!]
 \centering
 \begin{tabular}{| p{3.5cm}||c|c|p{8cm} |} 
 \hline
 \textbf{Name}& \textbf{[hours]}& \textbf{Speakers}& \textbf{Key Features}\\
 \hline
 HPSDB& \textasciitilde16& 308&Several types of speech, Parkinson's disease and healthy control.
 \citetlanguageresource{HPSDB}\\
 \hline
 LAPASDA& \textasciitilde7& \textasciitilde400&Read text and sustained sounds; dysphonia and healthy control, RBH scores.
 \citetlanguageresource{LAPASDA}\\
 \hline
 DEPISDA& \textasciitilde17& 400& Read text, spontaneous; depression and healthy control, BDI-II scores.
 \citetlanguageresource{DEPISDA}\\
 \hline 
 HuMenDisCo& n.a.& 90&Spontaneous; bipolar, schizophrenia and schizoaffective disorder and healthy control.
 \citetlanguageresource{HuMenDisCo}\\
 \hline 
 \textbf{Total approximately} & \textbf{40} & \textbf{1198} & \\\hline
 \end{tabular}
 \caption{Hungarian Pathological Speech Data Sets}
 \label{tab:patho}
\end{table*}

\subsection{Child-related Speech}
Table~\ref{tab:child} outlines data sets with child or child-directed speech (by adults). 
GABI (Child Language and Speech Database and Information Repository), the largest corpus containing recordings with 422 children/adolescents is currently under development.


\begin{table*}[ht!]
 \centering
 \begin{tabular}{| p{3.5cm}||c|c|p{8cm} |}
 \hline
 \textbf{Name}& \textbf{[hours]}& \textbf{Speakers}& \textbf{Key Features}\\
 \hline
 SPECO & \textasciitilde15& 72& 5-10 years old children, read speech and sustained segments. \citetlanguageresource{SPECO}\\
 \hline
 SPEECON-2& \textasciitilde10 & 50& 8 to 15-year-old children, home environment, read sentences, toy commands, etc. \citetlanguageresource{iskra2002speecon}\\
 \hline
 CHILDES& \textasciitilde13& 1& A boy between the ages 1;11 and 2;11, freeplay, home environment. \citetlanguageresource{reger2004hungarian}\\
 \hline
  Monyek \newline(Eng. HUKILC) & \textasciitilde20& 62& 4.5--5.5 ages old children, 20--30 minutes of talks. 
  \citetlanguageresource{orosz2014mlu}\\
 \hline
 TiniBea&\textasciitilde13& 18& 8 female, 10 male speakers of 16--18 years, similar protocol with BEA. \citetlanguageresource{tinibea}\\
 \hline 
 GABI& \textasciitilde50 & \textasciitilde100& 3-18 years old speakers, various speech types. \citetlanguageresource{bona2019gabi}\\
 \hline
 HIDS& \textasciitilde9& 68& Infant-directed speech, longitudinal; read text, semi-spontaneous 
 storytelling. \citetlanguageresource{kohari2023infant}\\
 \hline
 \textbf{Total approximately} & \textbf{117} & \textbf{353} & \\\hline
 \end{tabular}
 \caption{Hungarian Child-Related Data Sets}
 \label{tab:child}
\end{table*}

\subsection{Hungarian Dialects}
All data sets contain time-aligned narrow phonetic transcriptions (see Table~\ref{tab:dialect}). MNYHK (Hungarian Dialectal Talking Book Series) contains transcribed interviews or interview segments recorded between 1960-1964.
VIHk (Istv\'an V\"o\H{o} Talking Book) includes short interviews 
mostly from Transylvania (part of in Romania) in the 1960's. 
SzMNyHk (Talking Books of Hungarian Dialects in Slovakia) consists of interviews recorded recently in and around Nitra.
MoMa (Hungarian in Moldova) is composed of transcribed interviews with Moldavian Cs\'ang\'os starting from 1990. The Heged\H{u}s Archive  \citetlanguageresource{hege} is the first collection of Hungarian dialectal speech, recorded around the 1950s and digitized in 2005. 


\begin{table*}[ht!]
 \centering
 \begin{tabular}{| p{3.5cm}||c|c|p{8cm} |}
 \hline
 \textbf{Name}& \textbf{[hours]}& \textbf{Speakers}& \textbf{Key Features}\\
 \hline
 MNyHk & 13 & \textasciitilde110 & Interviews from 79 locations in and around Hungary. \citetlanguageresource{MNyHk}\\
 \hline
 VIHk & 1 & 18 & Interviews from 17 locations in Transylvania (Romania). \citetlanguageresource{VIHk}\\
 \hline
 SzMNyHk & 3 & 34 & Interviews in Hungarian dialects from Slovakia. \citetlanguageresource{SzMNyHk}\\
 \hline
 MoMa & 18 & \textasciitilde100 & Interviews in Moldavian (Romanian) Hungarian (Csángó). \citetlanguageresource{moma} \\
 \hline
 \textbf{Total approximately} & \textbf{35} & \textbf{262} & \\\hline
 \end{tabular}
 \caption{Hungarian Dialectal Speech Data Sets}
 \label{tab:dialect}
\end{table*}



 

\begin{figure}
\begin{center}
\includegraphics[width=3.5cm]{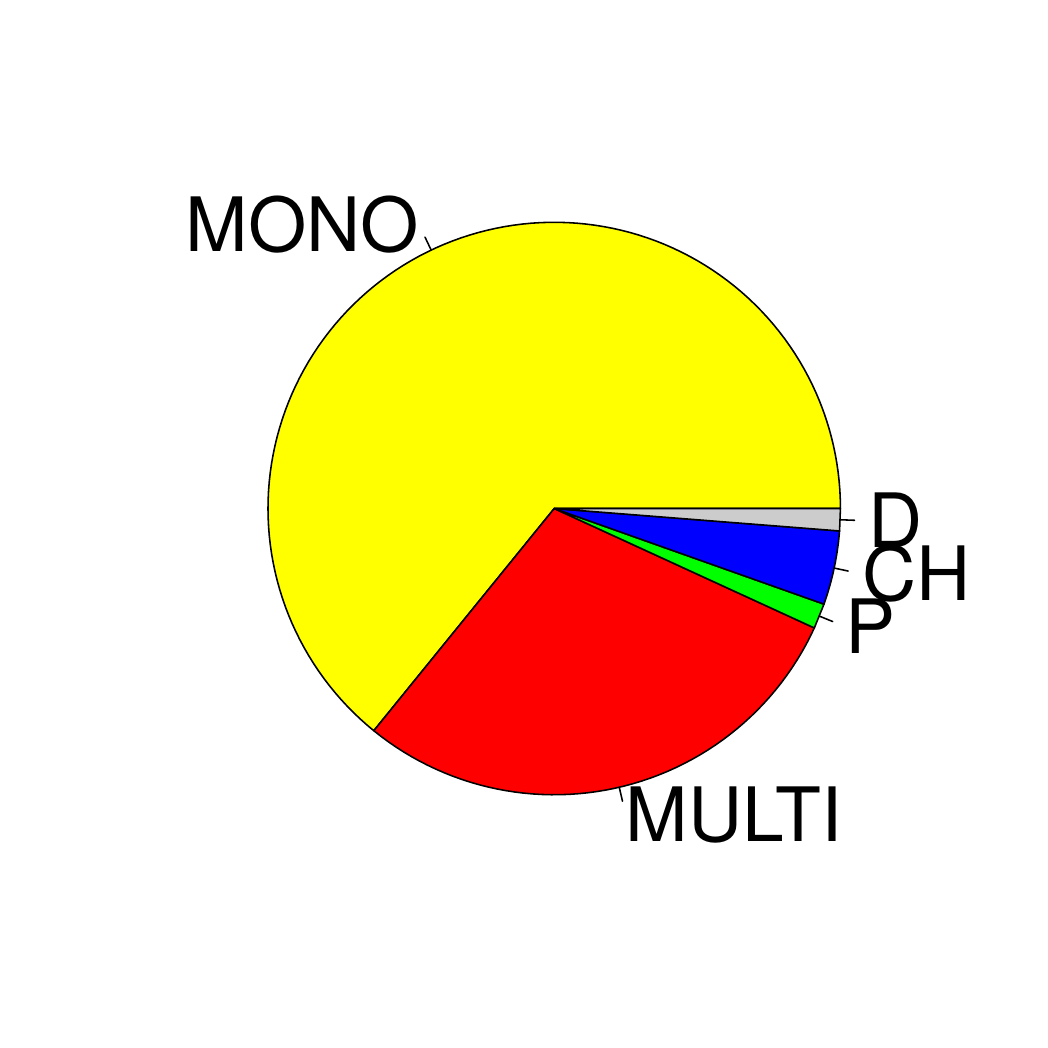}
\includegraphics[width=3.5cm]{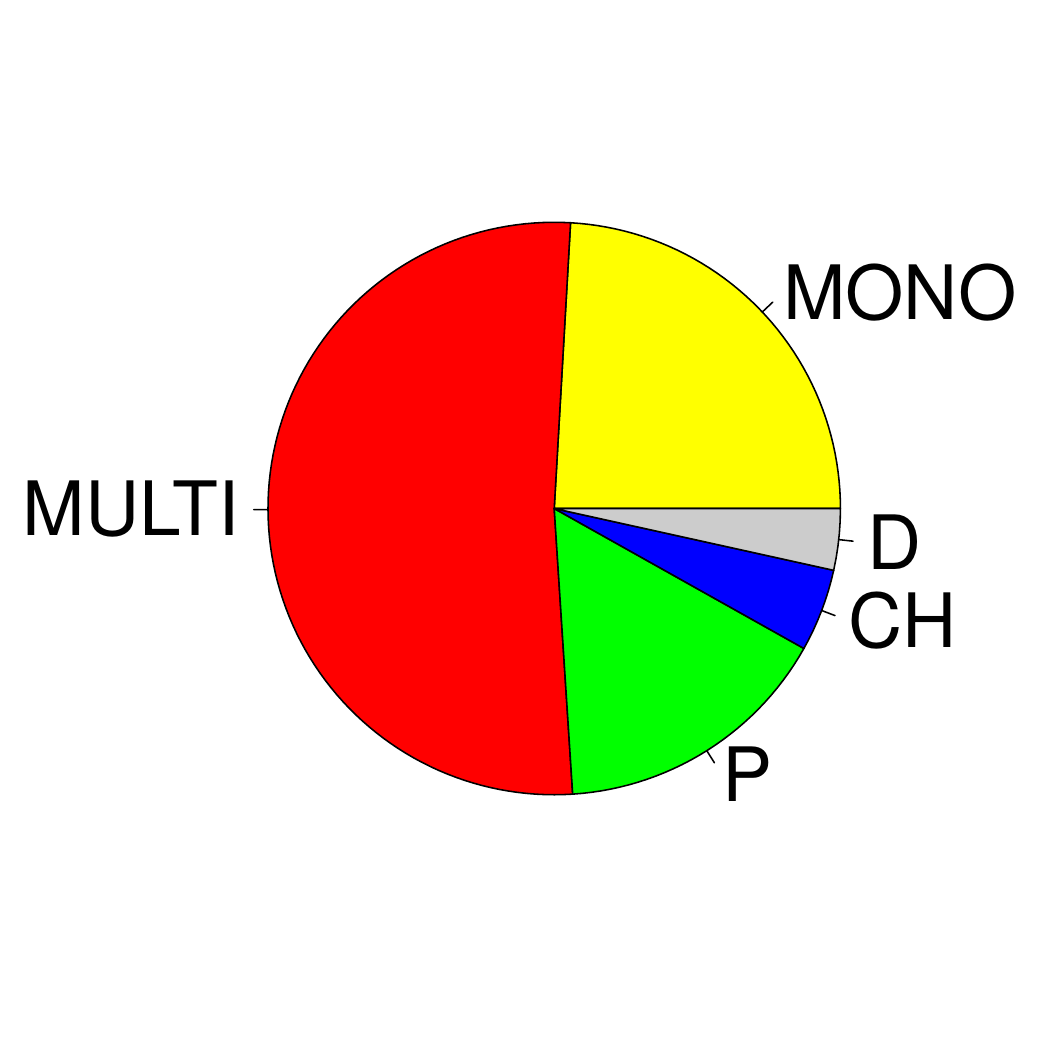}\vspace{-1cm}
\caption{Types of Hungarian data sets. Left: Length in hours, Right: Number of speakers. MONO: monolingual, MULTI: Hungarian from multilingual collections, P: pathological, CH: child-related, D: dialectal.}
\label{fig.1}
\end{center}
\end{figure}

\section{Discussion and Conclusions}


% ...
Although none of the data sets in the previous section is comparable in size and in variability (e.g. with LibriSpeech, \citetlanguageresource{librispeech} or AISHELL-2, \citetlanguageresource{aishell2}), summing all the Hungarian speech corpora, we still reach remarkable quantities of 3.000 hours and 8.000 speakers in our survey. According to Tables~1--5 and Figure~1, this hypothetical super data set is highly diverse and includes read and conversational speech, broadcast news, monologues, phone calls across diverse speakers (e.g., healthy adults, children, pathological and dialectal speakers)
. In that sense, spoken Hungarian may appear to be well-resourced. 
However, developing a speech recognition model based on the conglomerate of the previous data sets would be a challenging task. First, many of the speech corpora are difficult to access. 
Next, transcription format, style and methodologies are highly diverse. Transcription data is sometimes stored in a separate document without time-alignment, and some words (or entire texts) are annotated in their phonetic form instead of standard orthography. Annotation systems differ in how foreign words are handled, and abbreviations, numbers, hesitations, noises and punctuation marks vary across data sets without consistency even within the same corpus. Despite the impressive agglomerate numbers, the distribution of ``speech hours'' and speakers is rather uneven. One of the largest corpora (HuTongue) is collected only from 15 speakers, whereas corpora with high numbers of speakers are often tiny (MTBA, MRBA). This leads to issues about representativeness of the collected speech corpora in comparison to the diversity in the general population \cite{dogruoz-etal-2023-representativeness}.
In some data sets audio recording is not controlled sufficiently (e.g. BEKK, StaffTalk), or it is of studio quality (e.g., BEA, HuComTech). Read speech is overrepresented, and 
conversational speech is less easily accessible because of GDPR (General Data Protection Regulation)/privacy issues. All in all, potential training data does not align well with the foreseeable ASR applications (e.g., meeting transcription, e-mail dictation). Thus, spoken Hungarian can still be considered underresourced for Speech Technology. 
More specifically, larger scale data sets (e.g., SwitchBoard \citetlanguageresource{switchboard}), meetings or lectures are missing. Furthermore, the diversity of data sets/annotation approaches pose significant challenges if a unified foundational speech model was to be trained. 
Although new data sets arise and older ones are being processed further, the demand for larger, more aligned and more accessible transcribed speech data is growing rapidly. However, having an overview of existing resources is the first step for the unification of these data sets. We also hope that our survey will be beneficial for researchers who conduct research about Hungarian across disciplines. 


\section{Acknowledgement}
The authors in Hungary were supported partially by the NKFIH K143075, K135038 
and NKFIH-828-2/2021(MILAB) projects of the NRDI Fund.

\nocite{*}
\section{Bibliographical References}\label{sec:reference}

\bibliographystyle{lrec-coling2024-natbib}
\bibliography{lrec-coling2024-example}

\section{Language Resource References}
\label{lr:ref}
\bibliographystylelanguageresource{lrec-coling2024-natbib}
\bibliographylanguageresource{languageresource}

\end{document}